\pdfoutput=1

\documentclass[letterpaper]{article}

\usepackage{aaai23}
\usepackage{verbatim}
\usepackage{subcaption}
\usepackage{times}  
\usepackage{helvet}  
\usepackage{courier}  
\usepackage[hyphens]{url}  
\usepackage{graphicx} 
\usepackage{natbib}  
\usepackage{caption} 
\usepackage{algorithm}
\usepackage{algorithmic}

\usepackage{pifont}
\newcommand{\cmark}{\ding{51}}%
\newcommand{\xmark}{\ding{55}}%

\usepackage{amssymb}
\usepackage{amsmath}
\usepackage{booktabs}
\usepackage{multirow}

\usepackage{newfloat}
\usepackage{listings}
\DeclareCaptionStyle{ruled}{labelfont=normalfont,labelsep=colon,strut=off} 
\floatstyle{ruled}
\newfloat{listing}{tb}{lst}{}
\floatname{listing}{Listing}
\newcommand{\bs}{\boldsymbol}

\usepackage{xcolor}
\usepackage{colortbl}
\definecolor{demphcolor}{gray}{.5}
\newcommand{\demph}[1]{\textcolor{demphcolor}{#1}}

\usepackage{makecell}

\definecolor{gold}{rgb}{0.83, 0.69, 0.22}
\usepackage{xparse}
\iftrue  

\NewDocumentCommand{\heng}
{ mO{} }{\textcolor{red}{\textsuperscript{\textit{Heng}}\textsf{\textbf{\small[#1]}}}}
\NewDocumentCommand{\manling}
{ mO{} }{}
\NewDocumentCommand{\xudong}
{ mO{} }{\textcolor{cyan}{\textsuperscript{\textit{Xudong}}\textsf{\textbf{\small[#1]}}}}

\title{
Video Event Extraction via Tracking Visual States of Arguments
}

\author{
    Guang Yang,\textsuperscript{\rm *} 
    Manling Li,\textsuperscript{\rm \dag}
    Jiajie Zhang,\textsuperscript{\rm *}
    Xudong Lin,\textsuperscript{\rm \ddag}
    Shih-Fu Chang,\textsuperscript{\rm \ddag}
    Heng Ji \textsuperscript{\rm \dag}
}
\affiliations{
    \textsuperscript{\rm *}Tsinghua University,\\
    \textsuperscript{\rm \dag}University of Illinois at Urbana-Champaign,\\
    \textsuperscript{\rm \ddag}Columbia University\\
   yangg19@mails.tsinghua.edu.cn,
   hengji@illinois.edu
}

\usepackage{bibentry}

\begin{document}

\maketitle

\begin{abstract}

Video event extraction aims to detect salient events from a video and identify the arguments for each event as well as their semantic roles. 
Existing methods focus on capturing the overall visual scene of each frame, ignoring fine-grained argument-level information. 
Inspired by the definition of events as changes of  states, we propose a novel framework to detect video events by tracking the changes in the visual states of all involved arguments, which are expected to provide the most informative evidence for the extraction of video events. In order to capture the visual state changes of arguments, we decompose them into changes in pixels within objects, displacements of objects, and interactions among multiple arguments. We further propose Object State Embedding, Object Motion-aware Embedding and Argument Interaction Embedding to encode and track these changes respectively. Experiments on various video event extraction tasks demonstrate significant improvements compared to state-of-the-art models. In particular, on verb classification, we achieve 3.49\% absolute gains (19.53\% relative gains) in F\textsubscript{1}@5 on Video Situation Recognition.~\footnote[1]{Code and data resources are included in the supplementary materials and will be made publicly available for research purposes after publication. }
\end{abstract}


\section{Introduction}
\begin{figure}[ht]
    \centering
    \includegraphics[scale=0.33]{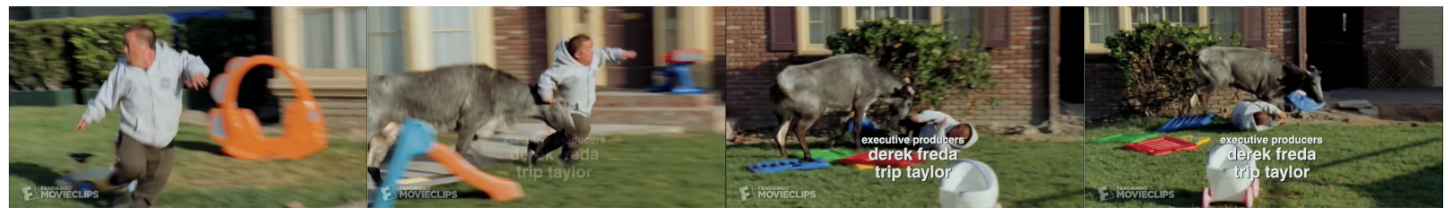}
    \caption{An example of a video event, with \textsc{knock} as the verb and \textit{man} as \textsc{target(arg\textsubscript{1})}, \textit{bull} as \textsc{agent(arg\textsubscript{0})}. 
    \manling{add table structure}
    }
    \label{img:task}
\end{figure}

The ability to comprehend a video requires a structured understanding of events, including what is happening, the objects involved, as well as their semantic roles. 
Situation Recognition~\cite{yatskar2016situation, pratt2020grounded} is a representative task for image event semantic structure understanding, aiming to classify an image to salient events (verbs)
and to predict entities (nouns) associated with semantic roles of each event. 
Video Situation Recognition~\cite{sadhu2021visual} extends the capacity of understanding event semantic structures from images to videos, as shown in Figure~\ref{img:task}. However, due to the high variation of temporal dynamics and visual features presented in videos, it remains to be a challenging and understudied task.

Traditional image event extraction~\cite{yatskar2016situation, pratt2020grounded,li2020cross,cho2022collaborative,wei2022rethinking} and video event extraction~\cite{medioni2001event,chen2021joint,sadhu2021visual} grasps the overall semantics of visual scenes, as a result of the difficulty in locating a specific visual area (i.e., a bounding box) to indicate the happening of an event. This is the case for images; however, additional temporal dynamics of videos provides another perspective to indicate the occurrences of events visually. 

Events, in linguistic form and interpretation, are telic, resultative, and indicate changes in states~\cite{dolling2008event}. 
Similarly, we argue that video events can, by nature, be represented by the visual state changes of all involved arguments. 
We further encode the \textbf{argument visual state changes} from three perspectives: (1) Single-frame state is characterized by in-bounding-box \textit{pixel changes} (such as the man's state changes from a running position to a lying position in Figure~\ref{img:task})  and \textit{displacements} of bounding boxes (such as the man moves from left to right in Figure~\ref{img:task}). 
Therefore, we propose \textbf{Object State Embedding} (OSE) by encoding both pixel changes and displacements. 
(2) The motion of objects across multiple frames, which we introduce \textbf{Object Motion-aware Embedding} (OME) to 
aggregate the changes of object states from different frames. 
(3) The interaction between objects, which we propose to learn an \textbf{Multi-Object Interaction Embedding} by encoding the interaction area,  which is able to capture the relative position of multiple objects and the pixel changes within intersection background. For example, the bounding boxes of \textit{man} and \textit{bull} are moving towards each other and then being apart, and the background pixels of \textit{grass} also indicate the  \textsc{knock down} event. 

On top of that, the information of multiple objects and their interactions is aggregated by an \textbf{Argument Interaction Encoder}, by \textit{contextualizing} objects with other objects within the same scene, and learning to focus on the objects and interactions that are more indictable. For example, the \textit{man} and the \textit{bull} in Figure~\ref{img:task} are the major objects compared to the \textit{orange ring} in the background. 
With this module design, our proposed model learns to rely on the salient objects that are possibly important argument roles for event detection.

As a result of this new paradigm of video event formulation, 
we classify a salient verb for each video by taking into account fine-grained states of arguments over time, as well as their interactions. 
The verb classification largely benefits from the joint modeling of multiple arguments, leading to 3.49\% absolute gains (19.53\% relative gains) in F\textsubscript{1}@5 according to the Video Situation Recognition validation set. Argument prediction enjoys a further boost due to the argument-level modeling, i.e., 3.97\% absolute gains on CIDEr.

Our contributions can be summarized as the follows:
\begin{itemize}
    \item We propose to formulate video events as the aggregation of visual state changes among multiple arguments. To the best of our knowledge, we are the first work to explore argument-level state changes in video event extraction. 
    \item We propose a new encoding framework by decomposing state changes into in-bounding-box pixel changes, bounding-box displacements, and multi-argument interaction.  
    \item Experiment results on two tasks, verb classification and argument prediction, demonstrate significant improvements compared to state-of-the-art models.
\end{itemize}   



\section{Methodology}
Our goal is to identify events (verbs) in a video clip, and to fill in the semantic roles of each predicted verb with entities (nouns). Our main idea is to introduce argument-level information by encoding the state changes of all arguments via Object State Embeddings and Object Motion-aware Embeddings, as well as their interactions through Multi-object Interaction Embeddings (Section~\ref{sec:embedding}). 
Afterwards, all objects are globally contextualized via an Argument Interaction Encoder (Section~\ref{sec:encoder}) to pay attention to objects that might be of relevance to main arguments for verb classification (Section~\ref{sec:verb}) and semantic role prediction (Section~\ref{sec:argument}). 

\subsection{Problem Formulation}

Given a video clip containing sampled frames $\{f_i\}_{i=0}^{F}$ with $F$ as the number of frames, the model is required to output a series of events $\{e\}$ occurring in the video. Each event is represented as a semantic structure
\begin{equation}
    e = \{v, 
    \langle r^0, a^0 \rangle, 
    \langle r^1, a^1 \rangle, 
    \cdots
    \},
\end{equation}
consisting a verb $v$ and multiple arguments $\{\langle r^k, a^k \rangle\}$ of the event. 
As defined in the event ontology, each verb $v$ is chosen from a pre-defined verb set $\mathcal{V}$, and has a pre-defined set of roles $\mathcal{R}(v)$. For example, the role set for the verb \textsc{knock} contains $\mathcal{R}(\textsc{knock})$ = $\{\textsc{agent}, \textsc{target}, \textsc{scene}, \dots\}$. $r^k \in \mathcal{R}(v)$ represents the $k$\textsuperscript{th} role of the verb $v$, and $a^k$ denotes the corresponding argument entity, which is described as a word or a phrase without a predefined set of words. 
For example, the event detected for the video clip in Figure~\ref{img:task} is $v=$ \textsc{knock}, and main arguments are $\langle$\textsc{agent(arg\textsubscript{0})}, \textit{gray bull}$\rangle$, 
$\langle$\textsc{target(arg\textsubscript{1})}, \textit{midget in gray hoodie}$\rangle$ 
and 
$\langle$\textsc{place(arg\textsubscript{scene})}, \textit{ground}$\rangle$
, etc. \manling{update}

\subsection{Object Visual State Tracking}
\label{sec:embedding}
\begin{figure}[ht]
    \centering
    \includegraphics[scale=0.4]{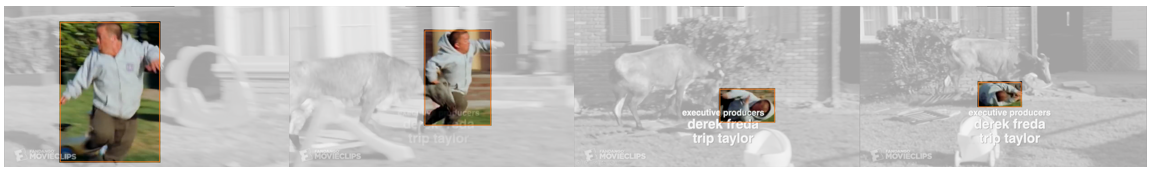}
    \caption{Bounding boxes of a man in a video.}
    \label{fig:man_motion}
\end{figure}

From the perspective of kinematics, the motion of an object can be decomposed into transitional and rotational movements, which correspond to the body shift and the rotation of the object, respectively.
Likewise, we apply this paradigm to videos and decompose the object motion into the \emph{displacement} of the bounding box and the \emph{pixel changes} within that bounding box.
As shown in Figure~\ref{fig:man_motion}, the displacement of the \textit{man} shows that he is moving from left to right and from near to far, while the pixel changes indicate that he is changing from standing to lying down. 
The motion of the \textit{man} can be well defined by the displacement of it bounding box and the pixel changes. Therefore, we detect objects in the video and propose object state embedding and object motion-aware embedding to encode the displacement and pixel changes of objects.

\paragraph{Object Detection and Tracking} 
We generate object tracklets using the state-of-the-art object tracking model VidVRD~\cite{gao2021video}. Taking a raw video as input, it outputs a list of object tracklets $\{o_j\}_{j=0}^{O}$, where $O$ is the number of recognized objects. 
Each object tracklet $o_j$ consists a list of its appearances in the video, and each apprearance is characterized by a time slot $\left[ t_{j}^{-}, t_{j}^{+} \right]$ and the coordinates of the left bottom and the right top corners of the bounding box 
\begin{equation}
    b_{ji} = \langle (b_{ji}^{0}, b_{ji}^{1}), (b_{ji}^{2}, b_{ji}^{3}) \rangle
\end{equation}
 at each time stamp inside the time slot. 
Take Figure~\ref{fig:man_motion} as an example. The object \textit{man} is detected on four frames with the bounding boxes extracted from each frame. 

\paragraph{Video Encoding Backbone} 
To utilize the superior power of 3D convolutional neural networks and improve computation efficiency,
our object state and motion-aware embedding use the state-of-the-art SlowFast model \cite{feichtenhofer2019slowfast} as the video encoder to get high-level grid features of raw videos. 
It includes a Slow pathway operating at a low frame rate to encode spatial semantics, and a Fast pathway operating at a high frame rate to capture motion at fine temporal resolution. 

The video encoder takes as input slow and fast sampled video frames $f^{\text{slow}}\in \mathbb{R}^{F_1\times W \times H \times 3}$ and $ f^{\text{fast}}\in \mathbb{R}^{F_2 \times W \times H \times 3}$, 
and outputs slow and fast grid features $\bs{g}^{\text{slow}}\in \mathbb{R}^{F_1\times W' \times H' \times d_1},
\bs{g}^{\text{fast}} \in \mathbb{R}^{F_2 \times W' \times H' \times d_2}$:
\begin{equation}
    [\bs{g}^{\text{slow}}, \bs{g}^{\text{fast}}] = \mathrm{SlowFast}(f^{\text{slow}}, f^{\text{fast}}),
\end{equation}
where 
$F_1$ and $F_2$ are the number of frames sampled from the video in slow and fast pathways, with the constraint of $F_1 < F_2$. Here, $W$ and $H$ are the width and height of raw video frames,
with $W'$ and $H'$ as the width and height of the processed grid features.
$d_1$ and $d_2$ are the number of channels of slow and fast grid features respectively.

\paragraph{Single Frame: Object State Embedding} 
First, in order to encode the states of objects in each frame, we introduce Object State Embedding. As shown in Figure~\ref{fig:man_motion}, we decompose the object state into in-bounding-box pixels for visual state tracking, as well as the position of the bounding box for displacement tracking. 

Given an object $o_j$ in a video frame $f_i$ and its appearance bounding box $b_{ji} = \langle (b_{ji}^{1}, b_{ji}^{2}), (b_{ji}^{3}, b_{ji}^{4}) \rangle$, we first project the corresponding bounding box to the grid feature space $ \hat b_{ji} = \langle (\hat b_{ji}^{1}, \hat b_{ji}^{2}), (\hat b_{ji}^{3}, \hat b_{ji}^{4}) \rangle$ by normalizing the coordinates.
To encode the state of the object $o_j$, we mask grid features outside the bounding box and exclusively focus on the object.


\begin{figure}
    \centering
    \includegraphics[scale=0.6]{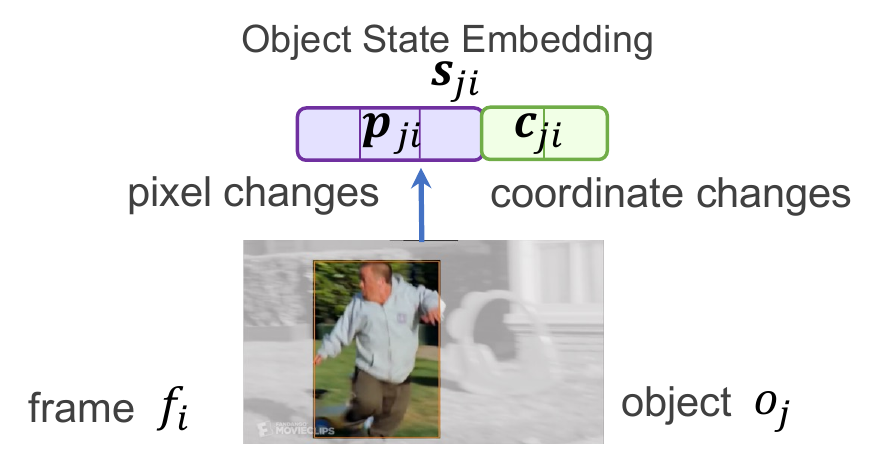}
    \caption{State embedding of the man in the first frame.}
    \label{fig:state}
\end{figure}

The visual state of an object $o_j$ is determined by its current status presented in the  frame $f_i$, such as the running status in Figure~\ref{fig:man_motion}. Thus, we adopt average pooling over the grid features within the bounding box: 
\begin{equation}
    \begin{aligned}
\bs{p}_{ji} & = \frac 1{(\hat b^{2}_{ji}-\hat b^{0}_{ji})(\hat b^{3}_{ji}-\hat b^{1}_{ji})} \sum_{x =\hat b^{0}_{ji}}^{\hat b^{2}_{ji}-1} \sum_{y = \hat b^{1}_{ji}}^{\hat b^{3}_{ji}-1} \bs{g}_{i}[x,y]. 
\end{aligned}
\end{equation}
where $(\hat b^{0}_{ji}, \hat b^{1}_{ji})$ and $(\hat b^{2}_{ji}, \hat b^{3}_{ji})$ represent the left bottom and the right top corners of the bounding box. $\bs{g}_{i}[x,y]$ represents the grid feature of the coordinate $\langle x,y \rangle$ in the grid feature space. The grid features $\bs{g}_{i}$ are from the slow pathway ($\bs{p}_{ji}^{\text{slow}} \in \mathbb{R}^{d_{1}}$) and the fast pathway ($\bs{p}_{ji}^{\text{fast}} \in \mathbb{R}^{d_{2}}$).  

To track the displacement of the object $o_j$, we encode its coordinates to track position changes: 
\begin{equation}
    \begin{aligned}
\bs{c}_{ji} & = \bs{W}_c\hat b_{ji},
\end{aligned}
\end{equation}
where $\bs{c}_{ji} \in \mathbb{R}^{d_c}$ is the coordinate embedding, with $d_c$ as its dimension. $\bs{W}_{c} \in \mathbb{R}^{d_{c} \times 4}$ is a positional embedding. 


As a result, the \textit{Object State Embedding} (OSE) $\bs{s}_{ji} \in \mathbb{R}^{d+d_c}$ is defined as the concatenation of pooled features $\bs{p}_{ji}$ and coordinate-based features $\bs{c}_{ji}$:
\[
\begin{aligned}
\bs{s}_{ji} & = [\bs{p}_{ji}, \bs{c}_{ji}]. 
\end{aligned}
\]

\begin{figure}[h]
    \centering
    \includegraphics[scale=0.4]{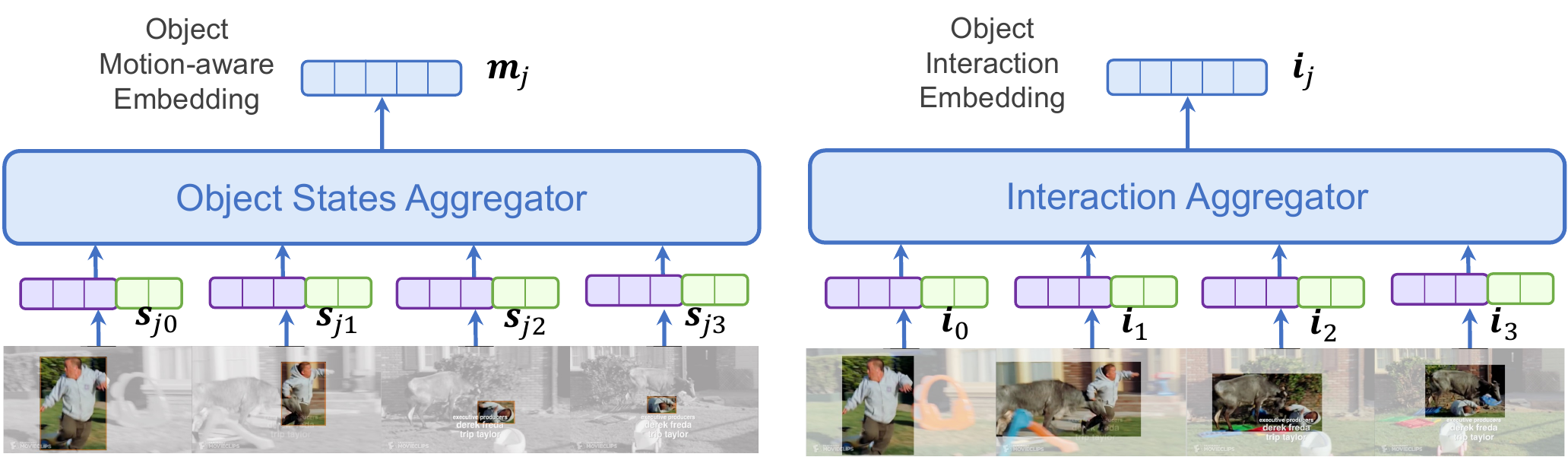}
    \caption{Object Motion-aware Embedding and Object Interaction Embedding.}
    \label{fig:motion_embedding}
\end{figure}

\paragraph{Multiple Frames: Object Motion-aware Embedding} 
To track the state changes of an object $o_j$ across multiple frames $\{f_i\}_{i=0}^{F}$ along the temporal dimension,
we feed object state embeddings of these frames into an \emph{object state aggregator}
to capture their differences, as shown in Figure~\ref{fig:motion_embedding}. 
It enables the model to capture the motion of the object $o_j$ when the video unfolds, so we call it \textit{Object Motion-aware Embedding} (OME) $\bs{m}_j$:  
\begin{equation}
    \bs{m}_j = \mathrm{StateAgg}(\{\bs{s}_{ji}\}_{i=0}^{F} ),
\end{equation}
where 
$\bs{m}_j \in \mathbb{R}^{d+d_c}$. 
The encoding is also done for both slow and fast pathways, i.e., $\bs{m}_j^{\text{slow}} \in \mathbb{R}^{d_1+d_c}, \bs{m}_j^{\text{fast}}\in \mathbb{R}^{d_2+d_c}$ respectively. 

Regarding the object state aggregator, we adopt LSTM~\cite{hochreiter1997long} as well as average pooling, and  perform further ablation studies on the choice of the aggregation operator. 

\paragraph{Multiple Objects: Object Interaction Embedding}

Apart from the motion of individual objects,
we further explore the joint state changes of multiple objects. 
As one example, the interaction between the \textit{man} and the \textit{bull} in Figure~\ref{fig:motion_embedding} contains the changes of relative positions by moving towards each other, then overlapping, and finally moving far apart. Also, the pixel changes of the interaction area between the \textit{man} and the \textit{bull}  provide a context, such as the building, the ground, the falling down circle ring, etc. 
We therefore propose {Object Interaction Embedding} to capture and track changes to the interaction area across objects.

To reveal an overall picture of how objects interact in a video frame $f_i$,
we first get the union bounding box $\hat{B}_i$ of these objects: 
\begin{equation}
    \hat{B}_i = \bigcup_{o_j \in \mathcal{O}_i} \hat{b}_{ji}
\end{equation}
where $\mathcal{O}_i$ is the set of objects in frame $f_i$ and $\cup$ computes the union of their bounding boxes.
Since the interaction area can also be characterized by the pixel changes and position changes, we use similar approach as OME to construct interaction states by the grid features within the interaction area: 
\begin{equation}
    \bs{p}'_{i} = \frac 1{(\hat B^{2}_{i}-\hat B^{0}_{i})(\hat B^{3}_{i}-\hat B^{1}_{i})} \sum_{x =\hat B^{0}_{i}}^{\hat B^{2}_{i}-1} \sum_{y = \hat B^{1}_{i}}^{\hat B^{3}_{i}-1} \bs{g}_{i}[x,y],
\end{equation}
and the coordinates showing the position of the area:  
\begin{equation}
\bs{c}'_{i} = \bs{W}_c\hat B_{i}.
\end{equation}
Similarly we represent the state of the interaction area of frame $f_i$ via a concatenation operation to gather information of both perspectives:
\begin{equation}
    \bs{i}_{i} = [\bs{p}'_{i}, \bs{c}'_{i}],
\end{equation}
where $\bs{i}_i \in \mathbb{R}^{d+d_c}$
After that, we employ an interaction aggregator to track the state changes of the interaction area as \textit{Object Interaction Embedding} (OIE) $\bs{i} \in \mathbb{R}^{d+d_c}$: 
\begin{equation}
    \bs{i} = \mathrm{InterAgg}(\{\bs{i}_{i}\}_{i=0}^{F}), 
\end{equation}
where $\mathrm{InterAgg}$ follows the same architecture as $\mathrm{StateAgg}$.
Notice that we get $\bs{i}^{\text{slow}} \in \mathbb{R}^{d_1+d_c}$ 
and $\bs{i}^{\text{fast}} \in \mathbb{R}^{d_2+d_c}$ through the slow and fast pathways, respectively. 

\subsection{Argument Interaction Encoder}
\label{sec:encoder}

Considering that not all state changes objects are relevant to the occurrence of the event, we contextualize each object in the video, 
and ask the model to learn to pay attention to the objects that are most relevant to the video and other objects. 
Therefore, we employ a Transformer layer \cite{vaswani2017attention} taking as input the overall embedding of entire video, the Object Motion-aware Embedding (OSE) $\{\bs{m}_{j}\}_{j=0}^{O}$ of candidate objects,
and the Object Interaction Embedding (OIE) $\bs{i}$ of the interaction area: 
\begin{equation}
    \bs{e} = \mathrm{Trans}(\ [\overline{\bs{g}}; \  \{\bs{m}_{j}\}_{j=0}^{O}; \  \bs{i}]\ ),
\end{equation}
where $\overline{\bs{g}} \in \mathbb{R}^{d}$ is the average-pooled features over all frames, and is used to represent the overall semantics of the video. 
The maximum number of objects $O$ is set as a hyperparameter, and the objects are ranked in terms of the confidence from the object detector.
It enables the output embedding to encode representative information to determine the event semantic structures, so we call it  \textit{event-aware visual embedding} $\bs{e}$.
For each embedding $\overline{\bs{g}}$, $\bs{m}_{j}$ and $\bs{i}$, we concatenate their embeddings from both slow and fast pathways, and further project $\bs{e} \in \mathbb{R}^{3d_1+3d_2+2d_c}$ to $\bs{e} \in \mathbb{R}^{d_1+d_2}$ using a linear layer. 


\begin{figure}
    \centering
    \includegraphics[scale=0.53]{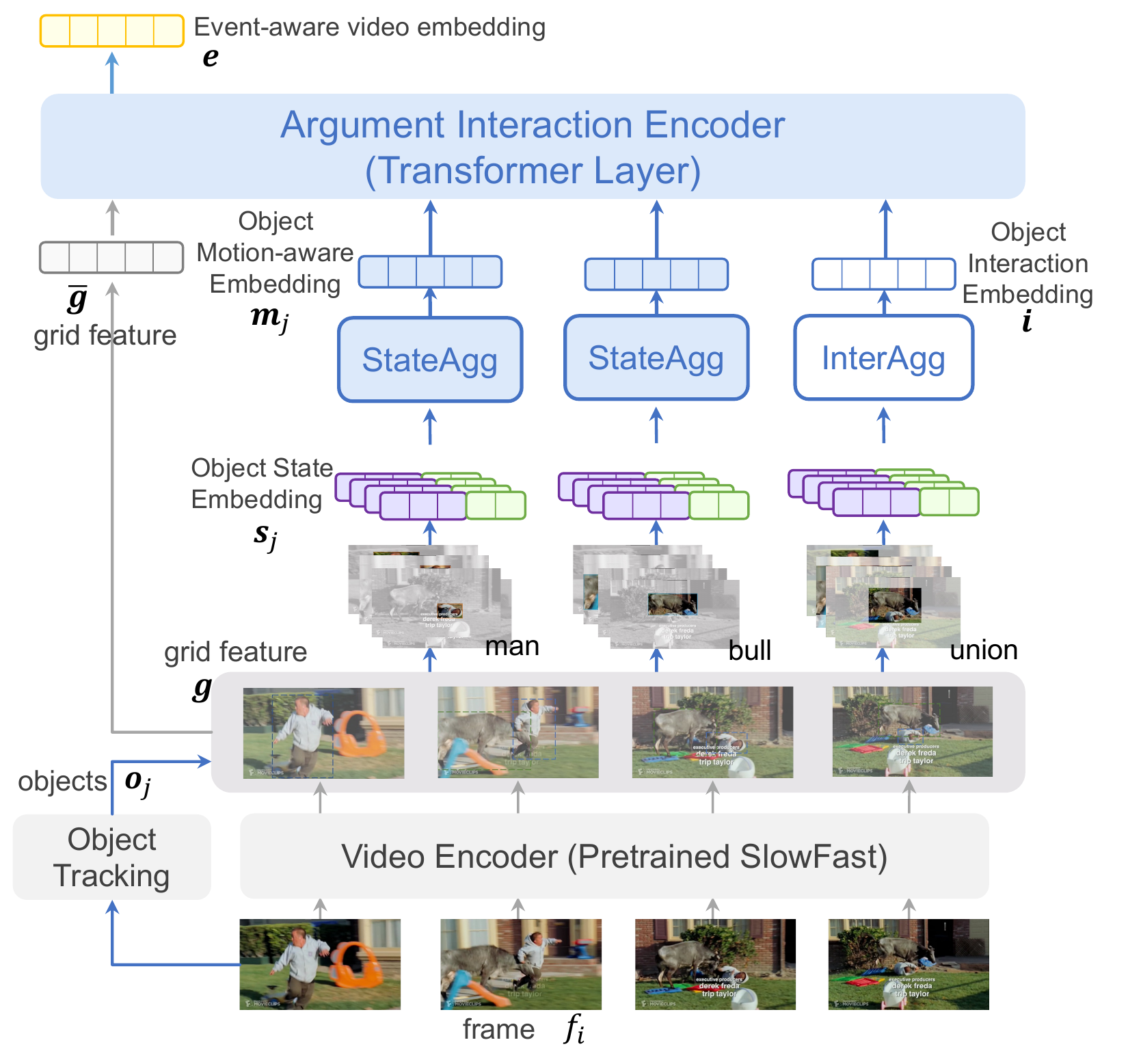}
    \caption{An overview of our event-aware visual embedding.}
    \label{fig:model}
\end{figure}

\subsection{Verb Classification}
\label{sec:verb}
Verb classification is a rudiment motion-related video understanding task to select verbs $\{v\}$ from a predefined event ontology\footnote{There are 2154 verb-senses defined in the event ontology of the video situation recognition dataset  VidSitu~\cite{sadhu2021visual}.}. Our event-aware visual embedding $\bs{e}$ captures the state changes of arguments, which is the most informative feature to identify events. 
As a result, we classify the video by 
feeding its event-aware visual embedding $\bs{e}$ into a classification layer
and train our model with a cross-entropy loss following~\cite{sadhu2021visual}.
The classification layer consists of two linear transformations 
with a ReLU activation layer in between:
\begin{equation}
    v = \mathrm{Softmax} \left( \bs{W}_2 \ \left[ \mathrm{ReLU} \ ( \bs{W}_1 \bs{e} + \bs{b}_1) \right] + \bs{b}_2 \right),
\end{equation}
which produces a ranked list of the predicted verb $v$ based on the $\mathrm{Softmax}$ probabilities. We set the bottleneck dimension to $\frac{1}{2} (d_1+d_2)$, namely,
$\bs{W}_1:\mathbb{R}^{d_1+d_2} \to \mathbb{R}^{\frac{1}{2} (d_1+d_2)}$
and $W_2: \mathbb{R}^{\frac{1}{2} (d_1+d_2)} \to \mathbb{R}^{|\mathcal{V}|}$,
where $\mathcal{V}$ is the set of candidate verbs defined in the event ontology.

\subsection{Semantic Role Prediction}
\label{sec:argument}
\begin{figure}[h]
    \centering
    \includegraphics[scale=0.43]{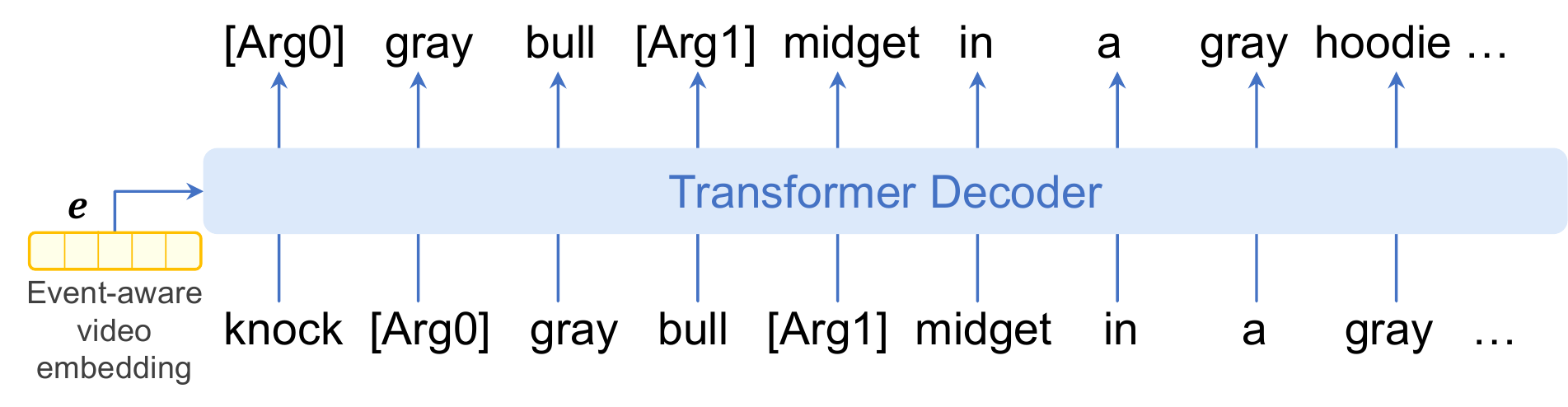}
    \caption{An overview of our model for semantic role prediction. 
    }
    \label{fig:arg_model}
\end{figure}

The most challenging part of event semantic structure extraction is to recognize the involved objects and their semantic roles. 
Our event-aware visual embedding $\bs{e}$ preserves fine-grained argument-level information, so we follow the architecture of state-of-the-art  model~\cite{sadhu2021visual} but replace its input SlowFast embedding with our event-aware visual embedding $\bs{e}$.      
In detail, the arguments are decoded in a sequence-to-sequence manner shown in Figure~\ref{fig:arg_model}, where the decoded sequence is the concatenation of the verb and its arguments 
``$v$ \texttt{[Arg$_{0}$]} $a^{0}$ \texttt{[Arg$_{1}$]} $a^{1}$ ...'', namely, the $k$\textsuperscript{th} argument $a^{k}$ is predicted by: 
\begin{equation}
    a^{k} = \mathrm{Decoder}(v \left[\text{\texttt{Arg}}_{0}\right] a^0 \left[\text{\texttt{Arg}}_{1}\right] a^1 \cdots \left[\text{\texttt{Arg}}_{k}\right]). 
\end{equation}
Our training objective is cross entropy loss between the generated sequence and the ground-truth sequence. We use teacher forcing strategy to obtain the verb $v$ during training. 






\section{Experiments}


\subsection{Dataset}

We evaluate our model on VidSitu \cite{sadhu2021visual}, the public video dataset providing extensive verb and argument structure annotations for more than 130k video clips. Each video clip is approximately two seconds in length with 10 verbs annotated in the evaluation split, and one verb annotated in the training split. There are 2154 verb-senses defined in the event ontology, and each verb has at least 3 semantic roles. 
Table~\ref{tab:statistics} shows the dataset statistics. 
Note that the testing split is hidden and preserved for a leaderboard\footnote{\url{https://leaderboard.allenai.org/vidsitu-verbs/submissions/public}}. 

\begin{table}[h]
\centering
\begin{tabular}{l |c c c c}
\toprule
 & Train & Valid & Test-Verb & Test-Role  \\
\midrule
\# Clip & 118,130 & 6,630 & 6,765 & 7,990  \\
\# Verb  & 118,130 & 66,300 & 67,650 & 79,900 \\
\# Role  & 118,130 & 19,890 & 20,295 & 23,970 \\
\bottomrule
\end{tabular}
\caption{Data statistics. 
}
\label{tab:statistics}
\end{table}




\begin{table*}
\setlength\tabcolsep{4pt}
\setlength\extrarowheight{1pt}
\centering
\begin{tabular}{l c|c c c c|c c c c}
\toprule
\multirow{2}{*}{Model} & \multirow{2}{*}{Kinetics} & \multicolumn{4}{c|}{Val} & \multicolumn{4}{c}{Test} \\
& & Acc@1 & Acc@5 & Rec@5 & F\textsubscript{1}@5 & Acc@1 & Acc@5 & Rec@5 & F\textsubscript{1}@5 \\
\midrule
TimeSformer & \cmark & 45.91 & 79.97 & 23.61 & 18.23 & - & - & - & - \\
\midrule
I3D$^\dagger$ & \xmark & 30.17 & 66.83 & 4.88 & 4.56 & 31.43 & 67.70 & 5.02 & 4.67 \\
SlowFast$^\dagger$ & \xmark & 32.64 & 69.22 & 6.11 & 5.61 &33.94 & 70.54 & 6.56 & 6.00 \\
I3D$^\dagger$ & \cmark & 29.65 & 60.77 & 18.21 & 14.01 & 29.87 & 59.10 & 19.54 & 14.68 \\
SlowFast$^\dagger$ & \cmark & 46.79 & 75.90 & 23.38 & 17.87 & 46.37 & 75.28 & 25.78 & 19.20 \\

\hline\rowcolor{gray!20}
\rowcolor{gray!20}
Ours (OSE-pixel\ \ \ \ \ \ \ \ \ \ +\ OME\ ) & \cmark & 52.75 & 83.88 & 28.44 & 21.24 & 52.14 & \textbf{83.84} & 30.66 & 22.45\\
\rowcolor{gray!20}
Ours (OSE-pixel/disp \ +\ OME\ ) & \cmark & 53.32 & \textbf{84.00} & 28.61 & 21.34 & 51.88& 83.55 & \textbf{30.83} & \textbf{22.52}\\
\rowcolor{gray!20}
Ours (OSE-pixel/disp \ +\ OME\ +\ OIE\ ) & \cmark & \textbf{53.36} & 83.94 & \textbf{28.72} & \textbf{21.40} & \textbf{52.39} & 83.47 & 30.74 & 22.47\\
\Xhline{2\arrayrulewidth}
\end{tabular}

\caption{\textbf{Results (\%) on verb classification.} ``Kinetics'' denotes whether the model is pre-trained on Kinetics-400 dataset. 
``OSE-pixel'' represents Object State Embedding with in-bounding-box features, and ``OSE-pixel/disp'' adds displacement capturing. ``OME'' stands for Object Motion-aware Embedding and ``OIE'' for Object Interaction Embedding. 
Results with $\dagger$ is the performance reported in VidSitu paper \cite{sadhu2021visual}. 
}
\label{tab:verb_class}

\end{table*}

\begin{table*}[h]
\centering
\setlength\tabcolsep{6pt}
\setlength\extrarowheight{1pt}
\begin{tabular}{m{5.5cm} |m{2.5cm}<{\centering} | m{2.5cm}<{\centering} | m{2.5cm}<{\centering} | m{2.5cm}<{\centering}}
\toprule
\multirow{2}{*}{Model} & CIDEr & CIDEr-Verb & CIDEr-Arg & ROUGE-L  \\
& Avg \quad Std & Avg \quad Std & Avg \quad Std & Avg \quad Std \\
\midrule
\demph{GPT2$^\dagger$} & \demph{34.67} & \demph{42.97} & \demph{34.45} & \demph{40.08} 
\\
\demph{I3D$^\dagger$} & \demph{47.06} & \demph{51.67} & \demph{42.76} & \demph{42.41} 
\\
\demph{SlowFast$^\dagger$} & \demph{45.52} & \demph{55.47} & \demph{42.82} & \demph{42.66} 
\\
SlowFast & 44.49 $\pm$2.30 & 51.73 $\pm$2.70 & 40.93 $\pm$2.42  & 40.83 $\pm$1.27 \\
\hline
\rowcolor{gray!20}
Ours (OSE-pixel\ \ \ \ \ \ \ \ \ \ +\ OME\ )  & 47.82 $\pm$2.12 & 54.51 $\pm$3.00 & 44.32 $\pm$2.45 & 40.91 $\pm$1.32 \\
\rowcolor{gray!20}
Ours (OSE-pixel/disp \ +\ OME\ )  & \textbf{48.46 $\pm$1.84} & \textbf{56.04 $\pm$2.12} & \textbf{44.60 $\pm$2.33} & \textbf{41.89 $\pm$1.12} \\
\rowcolor{gray!20}
Ours (OSE-pixel/disp \ +\ OME\ +\ OIE\ ) & 47.16 $\pm$1.71 & 53.96 $\pm$1.32 & 42.78 $\pm$2.74 & 40.86 $\pm$2.54 \\
\Xhline{2\arrayrulewidth}
\end{tabular}

\caption{\textbf{Results on semantic role prediction.} 
Results are the average (Avg) over 10 runs with standard deviation (Std) reported. The result with $\dagger$ is the single-run performance reported in VidSitu paper \cite{sadhu2021visual}, which is not a fair comparison due to the high variance of free-form generated argument names, so its color is dimmed.
}
\label{tab:sem_pred}

\end{table*}




\subsection{Evaluation Metrics}

\noindent
\textbf{Verb Classification}
Since each video clip in validation or test split is annotated with 10 verbs,
We predict top-$K$ verbs and calculate ranking-based metrics including Accuracy@$K$, Recall@$K$ and F\textsubscript{1}@$K$, and set $k=5$ in experiments following~\cite{sadhu2021visual}. Also, we report Accuracy@1 as an indicator of its precision in making predictions.

\noindent
\textbf{Semantic Role Prediction}
We follow~\cite{sadhu2021visual} to measure 
the prediction of \textsc{Arg$_{0}$}, \textsc{Arg$_{1}$}, \textsc{Arg$_{2}$}, \textsc{Arg\textsubscript{Loc}} and \textsc{Arg\textsubscript{Scene}}. Since the argument prediction is free-form generation, for every prediction of a semantic role,
we compute the CIDEr score~\cite{vedantam2015cider} by
treating the model prediction as candidate sentence 
and annotations as ground-truth descriptions.
Micro-averaged CIDEr score is computed over each individual prediction,
while macro-averaged CIDEr score is computed over every verb-sense (CIDEr-Verb) and 
over argument-types (CIDEr-Arg).
Also, 
we compute ROUGE-L \cite{lin2004rouge} for longest common subsequence
to show the completeness of generated arguments.

\subsection{Baselines}
\paragraph{State-of-the-art models}
We compare with the state-of-the-art model in~\cite{sadhu2021visual}, which contains two variants, 
including I3D~\cite{carreira2017quo} and SlowFast~\cite{feichtenhofer2019slowfast}.
For all baselines, we consider the variant with Non-Local blocks \cite{wang2018non},
which has been proved to be more effective according to VidSitu \cite{sadhu2021visual}. We provide their performance reported in the paper. 

We also include a strong transformer-based video model, TimeSformer~\cite{bertasius2021space} as an extra baseline to comprehensively assess the performance of our model. As it is not officially evaluated on VidSitu, we tried our best to search for a suitable hyper-parameter setting on this dataset. 

For semantic role prediction task, apart from reporting the performance of the state-of-the-art models in~\cite{sadhu2021visual}, we also include a text-only decoder GPT2~\cite{radford2019language} as a reference baseline since arguments are predicted via free-form generation. 

Unlike verb classification, semantic role prediction results have large variance between difference runs due to free-form generation of argument names. As a result, instead of only reporting performance of a single run, we perform 10 runs and report the average performance (Avg) as well as the standard deviation (Std), as shown in Table~\ref{tab:sem_pred}. 
For fair comparison, we re-run the state-of-the-art model\footnote{\url{https://github.com/TheShadow29/VidSitu}} using the best-performed SlowFast backbone for 10 times, and report its average and the standard deviation, which is comparable to the performance reported in its paper~\cite{sadhu2021visual}. 
We also include the original scores as a reference in Table~\ref{tab:sem_pred} but dim the color to indicate that they are not fair comparison.
Also, we are unable to report the performance on the hidden testing split due to the difficulty of submitting each model 10 times to leaderboard repeatedly, so the average performance is only reported on the validation split.



\paragraph{Ablation models} 
We include variants of our model to explore the effectiveness of our proposed mechanisms, including: (1) \textbf{OSE-pixel + OME} by only using Object State Embedding to track \textit{pixel} changes;
(2) \textbf{OSE-pixel/disp + OME} by using the full setting of Object State Embedding to track both \textbf{pixel} changes and \textit{displacements}; 
(3) \textbf{OSE-pixel/disp + OME + OIE} by using the full setting of Object State Embedding and adding Object Interaction Embedding as well.

Note that Object Motion-aware Embedding (OME) is not removable since it is a crucial component to connect multiple frames via a state aggregator, so we add additional experiments and discussions by exploring the different alternatives of the aggregator.

\subsection{Implementation Details}

\noindent
\paragraph{Verb Classification}
For verb classification task,
We trained our model for 10 epochs and 
report the model with highest validation F\textsubscript{1}@5 score.
The training took about 20 hour on 4 V100 GPUs,
comparable to the original SlowFast baseline. 
The frame sampling rate and dimensions of grid features are following SlowFast~\cite{feichtenhofer2019slowfast}. We keep $d_c = 128$ in our experiments. 
We set the maximum number of objects as $8$, and rank objects in terms of the detection confidence. 
The learning rate is chosen from $\{10^{-4}, 3 \times 10^{-5}\}$
and the batch size for training is set to 8.
We use the Adam optimizer with $\beta_1 = 0.9$, $\beta_2 = 0.99$ and $\epsilon = 10^{-8}$,
and no learning rate scheduler is applied to our training.
For all pre-trained parameters, we shrink their learning rate by 90\%,
which keeps the generalizability of the pre-trained encoder. 

\noindent
\paragraph{Semantic Role Prediction}
The visual event embedding of every video clip
is frozen after verb classification training.
During the training of semantic role prediction task, we only train the sequence-to-sequence model including the Transformer-based encoder and decoder.
The model is trained for 10 epochs and we report 
the performance with highest validation CIDEr.
The learning rate is fixed to $10^{-4}$
and the batch size is set to 8.
We also use the Adam optimizer with $\beta_1 = 0.9, \beta_2 = 0.99$ and $\epsilon=10^{-8}$.
Meanwhile, no learning rate scheduler is applied to our training.

\subsection{Quantitative Analysis}

\noindent
\textbf{Verb Classification}
As shown in Table \ref{tab:verb_class}, 
our model outperforms the state-of-the-art models on the leaderboard hidden testing set by 8.19\% $\sim$ 8.56\% absolute gains on Accuracy@5
and 4.88\% $\sim$ 5.05\% absolute gains on Recall@5. 
The overall F\textsubscript{1}@5 performance achieves 3.32\% absolute gains (17.3\% relative gains) on testing set and 3.53\% absolute gains (19.8\% relative gains) on the validation set. It shows the success of using argument state changes to determine events. 


\noindent
\paragraph{Semantic Role Prediction}
As shown in Table~\ref{tab:sem_pred}, our best performing model typically achieves 4\% absolute gains on CIDEr scores, proving the effectiveness of argument-level state tracking in identifying the role of the argument. Additionally, the standard deviation is reduced, demonstrating the robustness of fine-grained argument tracking in event structure parsing.

\noindent
\paragraph{Pixel Changes vs Displacements}
Comparison between OSE-pixel and OSE-pixel/disp proves that displacement is important in characterizing object states. It is especially effective in the task of semantic role prediction, showing that tracking the position changes of objects is very important for determining the motion of objects, thus being helpful for detecting semantic roles.

\noindent
\paragraph{Effect of Object Interaction Embedding (OIE)}
Verb classification performance is generally improved by capturing interactions between objects, since verbs, by their nature, describe the interactions between objects. However, it does not provide much assistance on argument role prediction, which focuses primarily on each individual argument.

\noindent
\paragraph{TimeSFormer Comparison}
Although our model uses a low-performing video encoding backbone SlowFast, the performance is higher on TimeSFormer, demonstrating the effectiveness of our event-aware visual embedding. 



\noindent
\paragraph{Effect of Object Number} 
As shown in Table~\ref{tab:objectnum}, 
the number of objects does not significantly affect performance, since the Argument Interaction Encoder calculates the attention across multiple objects and will focus on the objects that may play an important role in event detection. 
\begin{table}[ht]
\centering
\setlength\tabcolsep{8pt}
\begin{tabular}{l |c c c c}
\toprule
  & Acc@1 & Acc@5 & Rec@5 & F\textsubscript{1}@5  \\
\midrule
$O_{\text{max}}=2$ & 53.56	& 83.98	& 28.95	& 21.53 \\
$O_{\text{max}}=4$ & 53.56 &	84.10 &	28.51 &	21.29 \\
$O_{\text{max}}=8$ & 53.35 &	83.94 &	28.72 &	21.40 \\
\bottomrule
\end{tabular}
\caption{Results (\%) of different object numbers on verb classification. 
}
\end{table}

\noindent
\paragraph{State Aggregator Operation in OME} 
There is little difference in performance between various alternatives of object state aggregators, and average pooling is slightly more efficient, indicating that a few parameters are sufficient to encode changes of the state of an object. 

\begin{table}[h]
\centering
\setlength\tabcolsep{5pt}
\begin{tabular}{l |c c c c}
\toprule
  & Acc@1 & Acc@5 & Rec@5 & F\textsubscript{1}@5  \\
\midrule
LSTM & 53.23 &	83.76 &	28.57 &	21.30 \\
Average Pooling & 53.32	& 83.97 &	28.64 &	21.36 \\
\bottomrule
\end{tabular}
\caption{Results (\%) of different state aggregator operators on verb classification. 
}
\label{tab:objectnum}
\end{table}

\subsection{Qualitative Analysis}

\begin{figure}[h]
    \centering
    \includegraphics[width=\linewidth]{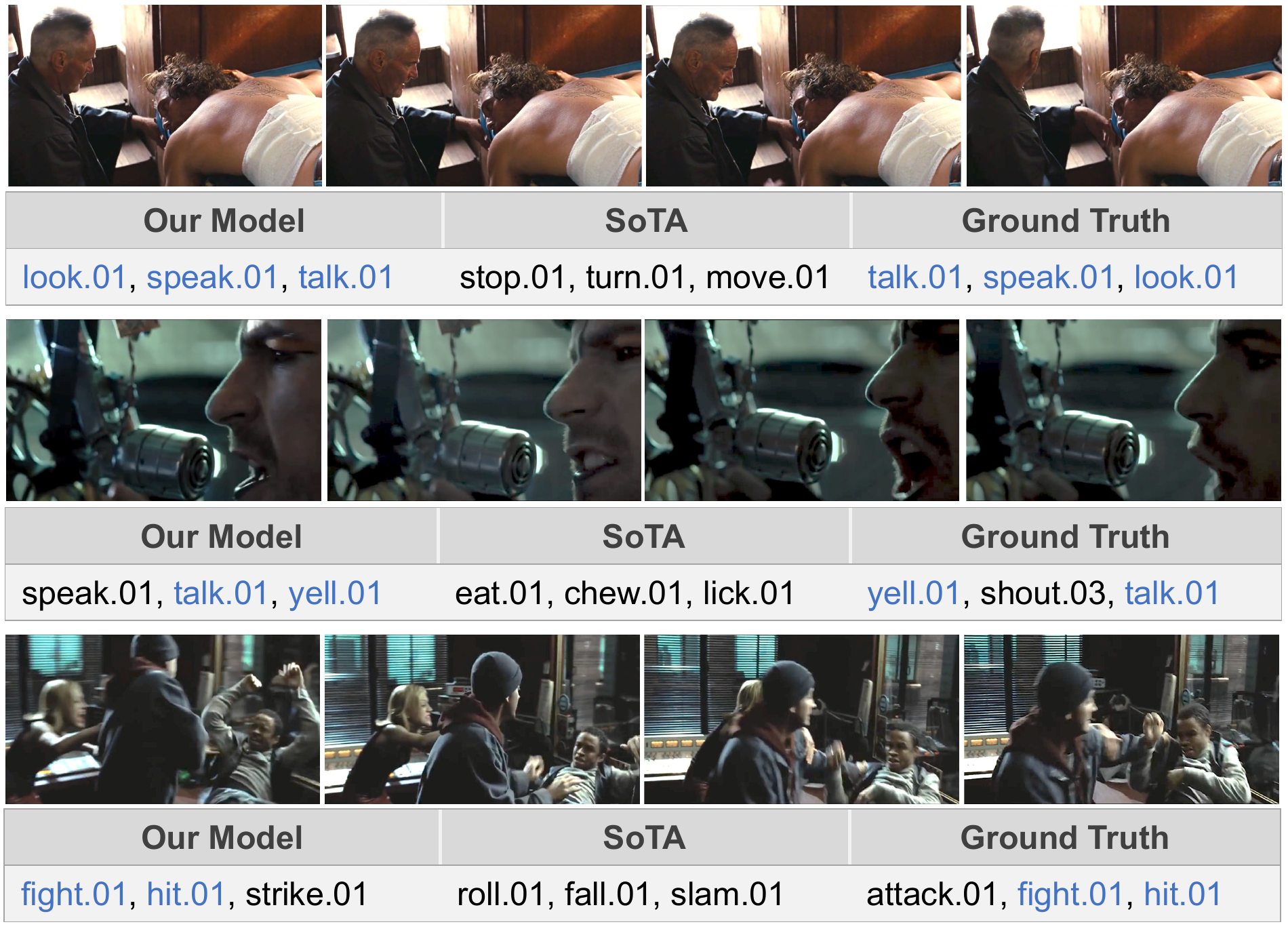}
    \caption{Result comparison between SoTA and our model. }
    \label{fig:case_study}
\end{figure}

Example results in Figure~\ref{fig:case_study} illustrate the successful use of argument state changes as a way of determining events. More examples and analysis are detailed in Appendix due to space limitation. 

According to the first example, two objects do not undergo much change in position or pixel, so our model is able to predict events corresponding to \textsc{look} and \textsc{talk}. However, the SoTA model fails to learn the correlation between object state changes and events, so it predicts events requiring state changes, such as \textsc{stop}, \textsc{turn} and \textsc{move}.
This example illustrates the reason why tracking changes is important to the detection of events. 

The second example illustrates how interactions contribute to the detection of events. 
A person's lips are moving as evidenced by pixel changes, so event \textsc{talk} or \textsc{chew} is largely relevant.
However, the event \textsc{chew} normally requires the interaction between two objects (a person and the food). Since food is not a participant and there is no interaction with food, our model correctly predict the event as \textsc{talk}. 
However, SoTA does not take into account fine-grained object-level interaction information, so it fails to predict the correct events but instead prefers \textsc{chew}. 

The third example demonstrates the important of encoding of pixel changes. 
It is evident from the position changes of the man on the right side of the video scene that he is moving downward, leading SoTA to prediction of \textsc{roll} and \textsc{fall}. However, his pixel changes indicate that his arm is waving emotionally, which is typically observed in \textsc{fighting} rather than \textsc{roll}. 
As a result, even when displacements match the event type, pixel changes remain extremely useful as an indicator of events.




\section{Related Work}

\noindent
\textbf{Event Extraction.} 
Extracting events from images/videos  \cite{yatskar2016situation, pratt2020grounded,sadhu2021visual}, texts \cite{ji2008refining,wang2019hmeae,liu2020event,LinACL2020}, or multimedia \cite{li2020cross,chen2021joint,wen2021resin,li2022clip} has attracted extensive research efforts. One of the key challenges in event extraction is to model the structural nature \cite{wang2019hmeae,li2020cross} of events and their associated argument roles. For example, in the text domain, \citet{wang2019hmeae} leverages modular networks to model the relationship between entities of different argument roles. However, to the best of our knowledge, we are the first to explicitly model the states of objects/entities and their relationships in the rich spatial-temporal dynamics of videos.

\noindent
\textbf{Video Understanding.} Video understanding generally requires the model to capture salient objects \cite{ikizler2010object,gu2018ava}, their motion \cite{wang2016temporal,carreira2017quo,Shou_2019_CVPR} and their interactions \cite{baradel2018object,gao2021video}. End-to-end models \cite{feichtenhofer2019slowfast,bertasius2021space,lin2020context} have shown their abilities to capture certain key information for classifying action in videos, where action is usually a single verb or a simple verb phrase~\cite{Kinetics}. Video event extraction \cite{sadhu2021visual} is a more complicated task than action recognition as it not only requires the model to understand the verb but also the interactions among multiple possible argument roles. \citet{baradel2018object} proposed to learn a reasoning model based on object features for action recognition. However, we first propose to explicitly model the object state, their changes and the interaction of objects for video event extraction.

\section{Conclusions and Future Work}

In this work, built upon modern 3D convolutional neural networks,
we construct the first event-aware video encoding neural network by tracking arguments visual states, 
which enables the model to understand event semantic structures. 
On both verb classification and semantic role prediction tasks in VidSitu,
our model achieved a new state of the art, 
outperforming state-of-the-art 3D convolutional neural networks. 
The promising future of our event-aware video embedding excites us, 
as it is applicable to a wide range of event-related video tasks, including action classification, video description, etc. 
In addition to adapting our model to more tasks, we will also work on tracking argument states in pre-training stages.

\bibliography{aaai23}

\appendix

\section{Implementation Details}

\subsection{Semantic Role Prediction}
To clarify, we freeze the model trained for verb classification task
and get $\bs{e}, \bs{m}_{j}, \bs{i}$ for every clip.
After that, we follow~\cite{sadhu2021visual} to contextualize the video clip with other clips in the same long video, which has been proved to have better performance. 
In order to combine the event embedding with object and interaction embeddings,
for each video clip $i\in\{1, 2, 3, 4, 5\}$, 
we define:

\[
\bs{e}' = \left[\bs{e}, \frac 1{|\mathcal{O}_i|+1} \left( \sum_{o_j \in \mathcal{O}_i} \bs{m}_{j} + \bs{i}\right)\right],
\]

\noindent where $\mathcal{O}_i$ is the set of objects in video clip $i$ with top-$K$ confidence.
Since a video is split into 5 clips,
to get a joint understanding about the story,
$e_1, e_2, e_3, e_4, e_5$ are passed into a Transformer \cite{vaswani2017attention} encoder,
getting a contextualized event embedding sequence $e'_1, e'_2, e'_3, e'_4, e'_5$.
For every single clip,
we follow~\cite{sadhu2021visual} to feed the contextualized event embedding $e'_i$ 
into a TransFormer decoder to generate a sequence of arguments and roles.

To point out, for models without the OIE,
we ignore the $\bs{i}$ term when computing the average.
Therefore,
\[
\bs{e}' = \left[\bs{e}, \frac 1{|\mathcal{O}_i|} \sum_{o_j \in \mathcal{O}_i} \bs{m}_{j}\right].
\]

To reproduce our results on semantic role prediction,
simply set the random seeds to 
\[
\{17, 33, 66, 74, 98, 137, 265, 314, 590, 788\}
\]
in our codes and get the mean and standard variation.
Table \ref{tab:all_results} compares SoTA and our best model
with CIDEr metric of every single run.

\begin{table}[h]
    \centering
    \setlength\tabcolsep{10pt}
    \begin{tabular}{l|c|c}
        \toprule
        Random seed & SlowFast &  Ours
        \\
        \midrule
        17 & 45.92 &50.41\\
        33 & 46.28&50.66\\
        66 & 41.92&47.61\\
        74 & 48.43&49.86\\
        98 & 46.97&47.80\\
        137 & 43.75&46.77\\
        265 & 44.28&49.60\\
        314 & 43.03&49.16\\
        590 &41.73 &47.98\\
        788 & 42.60&44.72\\
        \midrule 
        AVG & 44.49 & 48.46 \\
        STD & 2.30 & 1.84 \\
        \bottomrule
    \end{tabular}
    \caption{CIDEr metric of SoTA and our best model (OSE-pixel/disp + OME) on Semantic Role Prediciton.
    We show results of every single run.}
    \label{tab:all_results}
\end{table}



\section{Result Visualization}

\subsection{Argument Interaction Encoder Visualization}

To show the effectiveness of capturing argument-level information,
we visualize the attention map of the argument interaction encoder.
We stick to the example of the ``bull-knock-man'' event in the main paper.

\begin{figure*}[t]
    \centering
    \begin{subfigure}[b]{0.48\textwidth}
    \includegraphics[width=\textwidth]{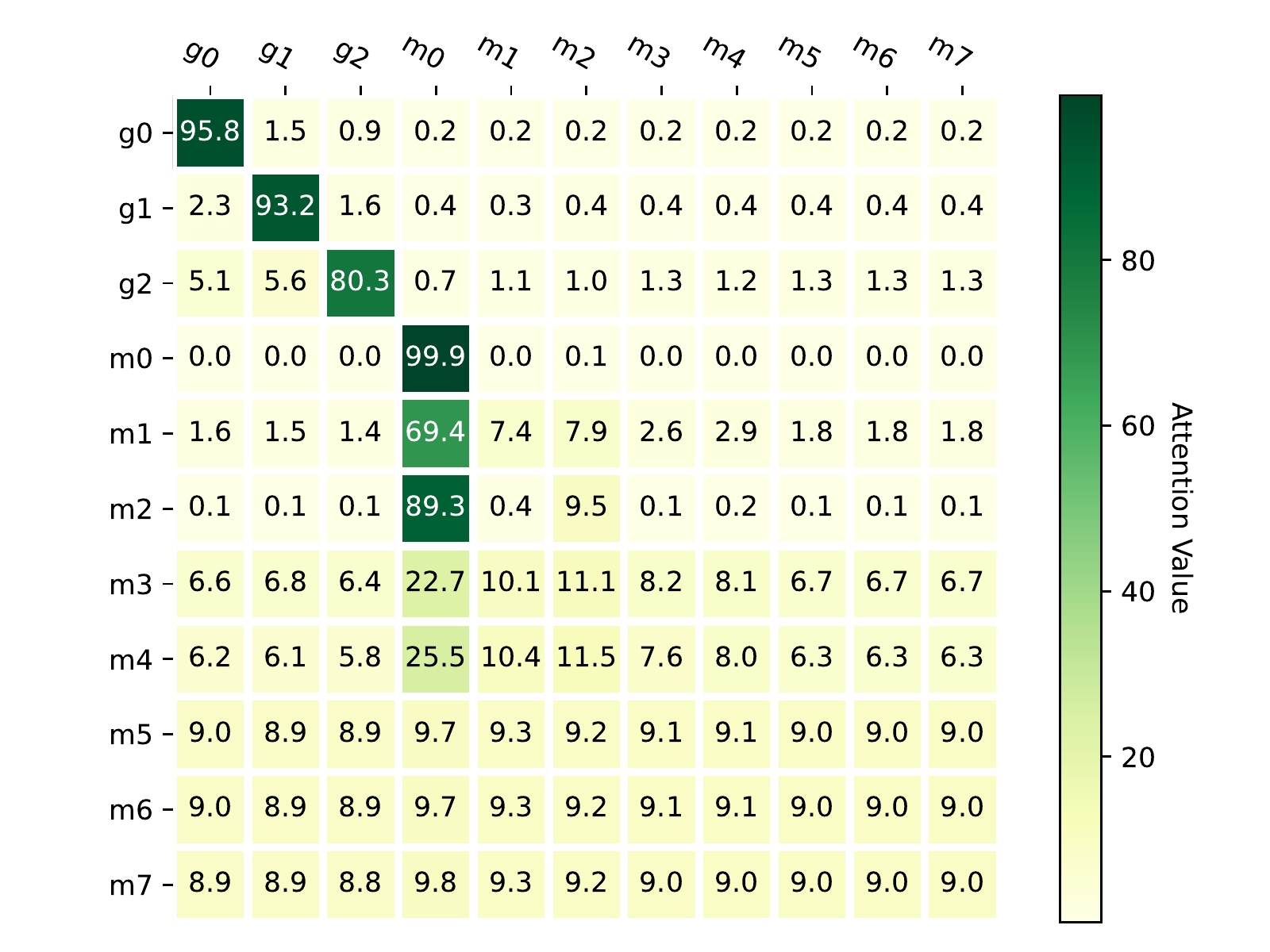}
    \caption{Argument attention visualization for our ``OSE-pixel/disp + OME'' model.}
    \label{fig:atten}
    \end{subfigure}
    \hfill
    \begin{subfigure}[b]{0.48\textwidth}
    \includegraphics[width=\textwidth]{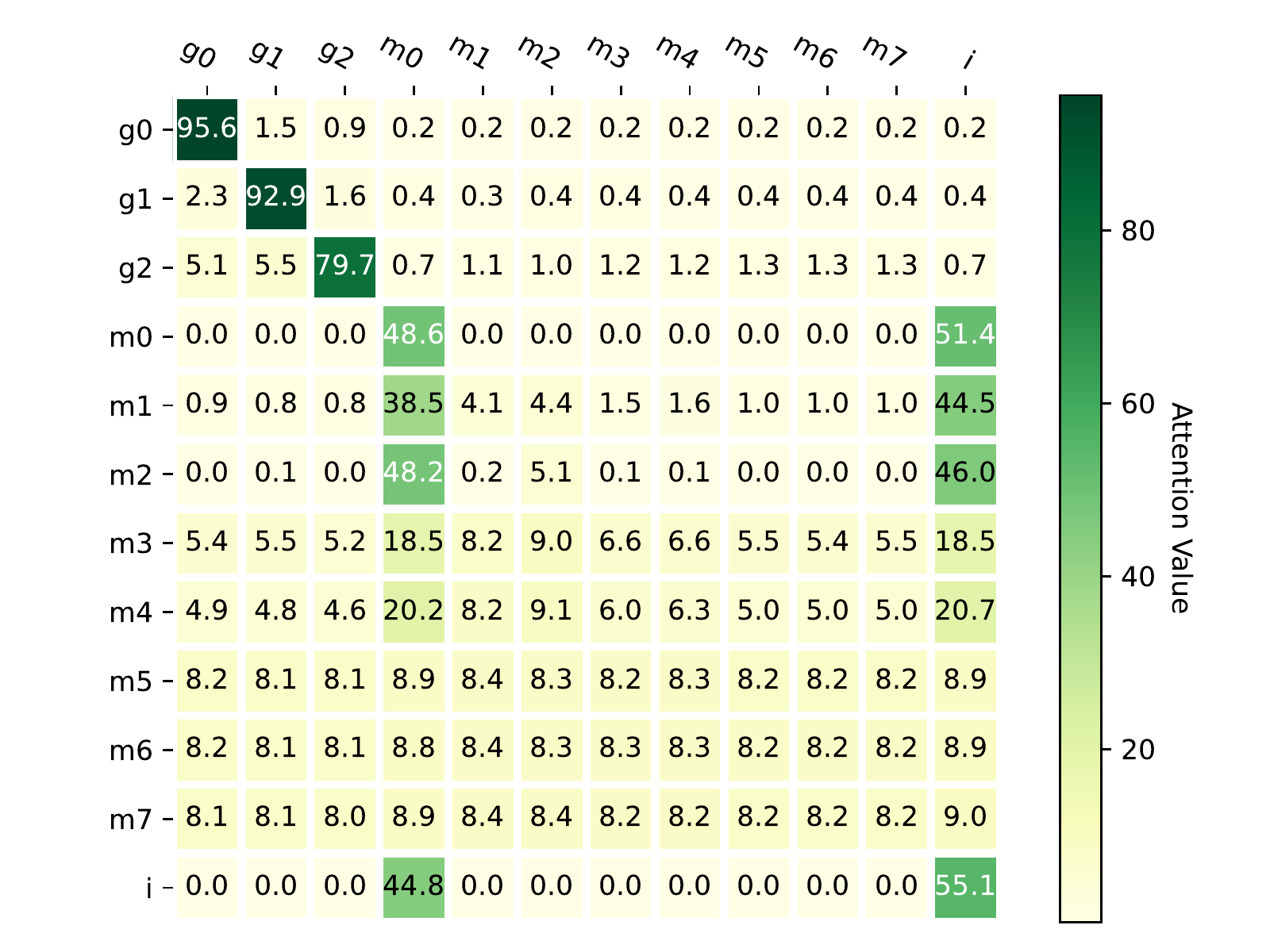}
    \caption{Argument attention visualization for our ``OSE-pixel/disp + OME + OIE'' model.}
    \label{fig:atten_inter}
    \end{subfigure}
    \caption{Argument attention visualization.}
\end{figure*}

In our ``OSE-pixe/disp + OME'' model,
the input sequence of the argument interaction encoder is: 
\[
\bar{\bs{g}}_0; \bar{\bs{g}}_1; \bar{\bs{g}}_2; \bs{m}_0; \bs{m}_1; \cdots; \bs{m}_8.
\]
We visualize the attention heat map between them in Figure \ref{fig:atten}.

\begin{figure*}[!ht]
    \centering
    \includegraphics[scale=0.35]{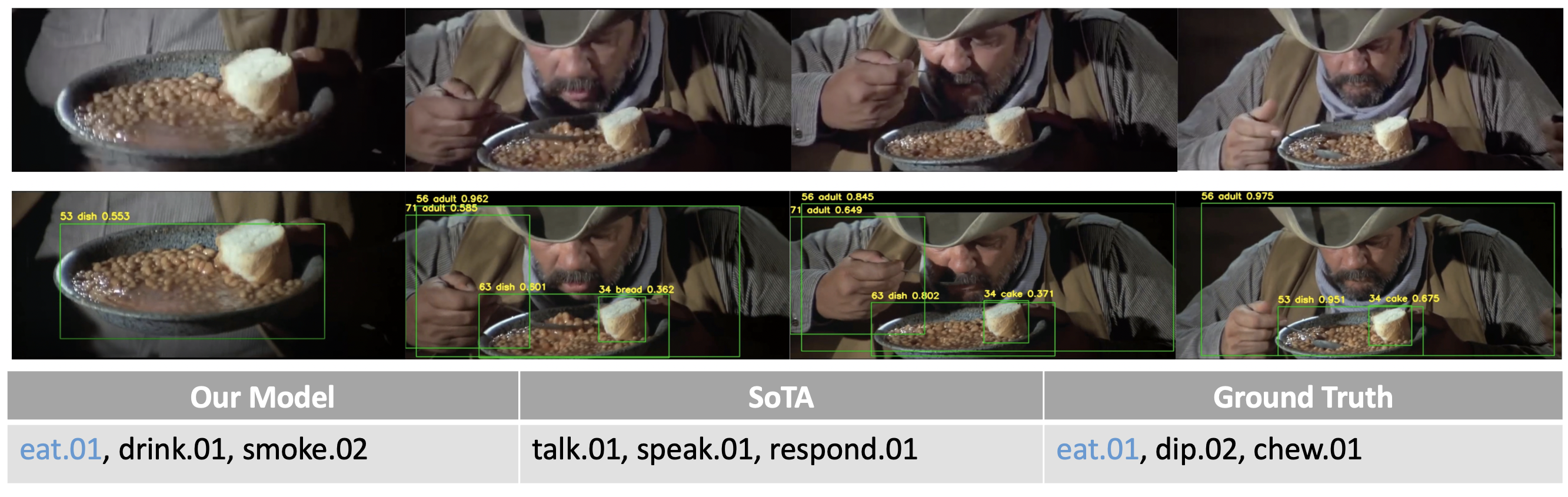}
    \caption{Detected Objects and Their Effectiveness.}
    \label{fig:eat_bread}
\end{figure*}

From the heat map in Figure~\ref{fig:atten}, we can observe that 
the interactions between objects are explicitly captured,
which not only facilitates the classification,
but also assists in the supervision of the video backbone SlowFast \cite{feichtenhofer2019slowfast} model.
In this approach,
multiple objects are incorporated into 
the event-aware video embedding with different weights,
thereby selecting the objects that are significantly involved in the event.
Meanwhile, the backbone model would be trained to give a relative close regional feature
if the objects are interacted with each other.

In our ``OSE-pixe/disp + OME + OIE'' model,
the object interaction embedding is appended into the input sequence: 
\[
\bar{\bs{g}}_0; \bar{\bs{g}}_1; \bar{\bs{g}}_2; \bs{m}_0; \bs{m}_1; \cdots; \bs{m}_8; \bs{i},
\]
and the attention heat map is shown in Figure \ref{fig:atten_inter}.

The object interaction embedding serves as an object-to-object joint,
by which the object embeddings are injected with inter-bounding-box information.
As we can see in the heat map in Figure~\ref{fig:atten_inter},
the object motion-aware contributes more significantly to the event-aware video embedding, compared to object interaction embedding, showing that tracking the state changes of object is more crucial to determine the type of events.

\subsection{Visualization of Object Detection and Tracking}

The object detection and tracking is the foundation of our model
and the advances in their capability enables us to achieve satisfactory performance. 
To show the role of object tracking method in our method,
we visualize the bounding boxes in Figure~\ref{fig:eat_bread}.

In this example,
SoTA model mistakenly give the answer ``speak'' because 
it only focuses on the movement
of the man's mouth,
but ignores the exsitense of objects in the background such as the bread.

In our model,
the dish, the man and the bread are successfully captured by the object tracker.
Those objects help our model to give the correct answer ``eat'' rather than ``speak''.


\end{document}